\documentclass[10pt,twocolumn,letterpaper]{article}

\usepackage{iccv}
\usepackage{times}
\usepackage{epsfig}
\usepackage{graphicx}
\usepackage{amsmath}
\usepackage{amssymb}
\usepackage{makecell}
\usepackage{multirow}
\usepackage{xcolor}
\usepackage{pifont}
\usepackage[font=small]{caption}
\usepackage{paralist}
\usepackage[accsupp]{axessibility}

\newcommand{\abc}{}
\newcommand{\jimmy}{}
\newcommand{\jimmyy}{}
\newcommand{\yty}{}

\usepackage[breaklinks=true,bookmarks=false]{hyperref}

\iccvfinalcopy 

\def\iccvPaperID{****} 

\ificcvfinal\fi

\begin{document}

\title{Scalable Video Object Segmentation with Simplified Framework}


\author{Qiangqiang Wu$^{1}$ \hspace{5mm} Tianyu Yang$^{2}\thanks{Corresponding Author}$ \hspace{5mm} Wei Wu$^{1}$ \hspace{5mm} Antoni B. Chan$^{1}$ \\
$^{1}${Department of Computer Science, City University of Hong Kong}\\
$^{2}${International Digital Economy Academy} \\
{\tt\small \{qiangqwu2-c, weiwu56-c-c\}@my.cityu.edu.hk, tianyu-yang@outlook.com}\\ {\tt\small abchan@cityu.edu.hk}
}

\maketitle
\ificcvfinal\thispagestyle{empty}\fi

\begin{abstract}
   The current popular methods for video object segmentation (VOS)
implement feature matching through several hand-crafted modules that separately perform  feature extraction and matching. However, the above \yty{hand-crafted} designs empirically cause insufficient target interaction, thus limiting the dynamic target-aware feature learning in VOS. To tackle these limitations, this paper presents a scalable Simplified VOS (SimVOS) framework to perform joint feature extraction and matching by leveraging a single transformer backbone. Specifically, SimVOS employs a scalable ViT backbone for simultaneous feature extraction and matching between query and reference features. This design enables SimVOS to learn better target-ware  features for accurate mask prediction. More importantly, SimVOS could directly apply well-pretrained ViT backbones (e.g., MAE \cite{mae}) for VOS, which bridges the gap between VOS and large-scale self-supervised pre-training. To achieve \yty{a} better performance-speed trade-off, we further explore within-frame attention and propose a new token refinement module to improve the running speed and save computational cost. Experimentally, our SimVOS achieves state-of-the-art results on popular video object segmentation benchmarks, i.e., \jimmyy{DAVIS-2017 (88.0\% $\mathcal{J}\&\mathcal{F}$), DAVIS-2016 (92.9\% $\mathcal{J}\&\mathcal{F}$) and YouTube-VOS 2019 (84.2\% $\mathcal{J}\&\mathcal{F}$), without applying any \yty{synthetic video} 
or BL30K pre-training used in previous VOS approaches. 
}
\end{abstract}

\section{Introduction}
Video Object Segmentation (VOS) is an essential and fundamental computer vision tasks in video analysis \cite{STM,pul,dsnet,AOT,dropmae,xia_wu_its} and scene understanding \cite{SOEnet,Coverage,c2mot,ljli,Blitznet,noisecount}. In this paper, we focus on the semi-supervised VOS task, which aims to segment and track the objects of interest in each frame of a video, using only the mask annotation of the target in the first frame as given. The key challenges in VOS mainly lie in two aspects: 1) how to effectively distinguish the target from the background distractors; 2) how to accurately match the target across various frames in a video.

In the past few years, modern matching-based VOS approaches have gained much attention due to their promising performance on popular VOS benchmarks \cite{youtubevos,davis16,davis17}. The typical method STM \cite{STM}  and its following works \cite{STCN,AOT,XMEM} mainly use several customized modules to perform semi-supervised VOS, including  feature extraction, target matching and mask prediction modules. The whole mask prediction process in these approaches can be divided into two sequential steps: 1) feature extraction on the previous frames (i.e., memory frames) and the new incoming frame (i.e., search frame); and 2) target matching in the search frame, which is commonly achieved by calculating per-pixel matching between the memory frames' embeddings and the search frame embedding. 

Despite the favorable performance achieved by the above matching-based approaches, the separated feature extraction and matching modules used in these methods still have several limitations. Firstly, the separate schema is unable to extract dynamic target-aware features, since there is no interaction between the memory and search frame embeddings during the feature extraction. In this way, the feature extraction module is treated as the fixed feature extractor after offline training and thus cannot handle objects with large appearance variations in different frames of a video. Secondly, the matching module built upon the extracted features needs to be carefully designed to perform sufficient \jimmy{interaction between query and memory features}. 
Recent works (e.g., FEELVOS \cite{Feelvos} and CFBI \cite{CFBI}) explore to use local and global matching mechanisms. However, their performance is still degraded due to the limited expressive power of the extracted fixed features.

 \begin{figure}
\begin{center}
   \includegraphics[width=1.0\linewidth]{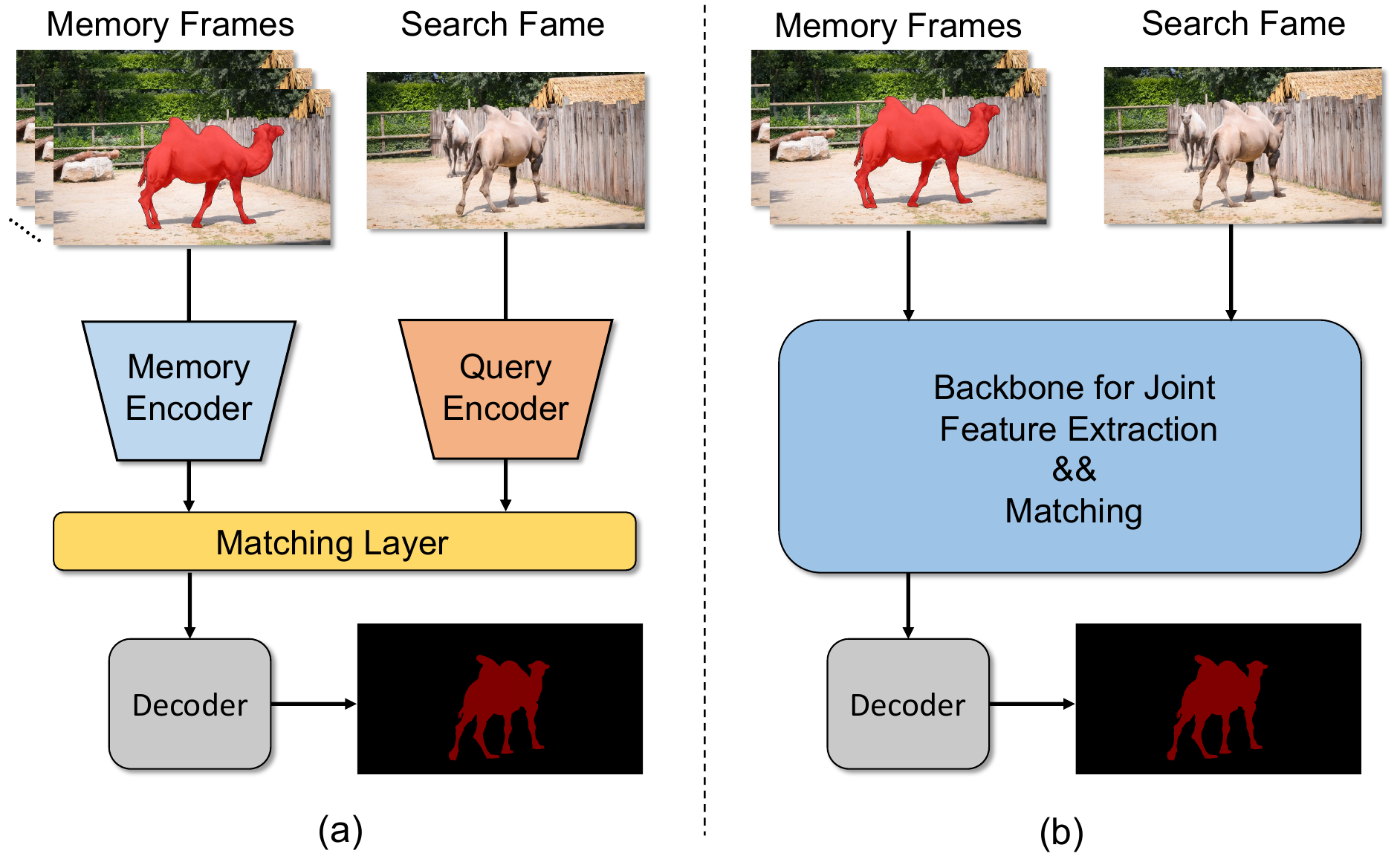} 
\end{center}
 \caption{A comparison between the pipelines of (a) traditional VOS approaches \cite{STM,STCN}, and (b) our proposed SimVOS. The previous approaches predict segmentation masks by leveraging the \jimmy{customized separate} feature extraction and matching modules. Our SimVOS removes \jimmy{hand-crafted} designs and employs a unified transformer backbone for joint feature extraction and matching, which provides a simplified framework for accurate VOS.}
 \label{archi_compare}
\end{figure}

To solve the aforementioned problems, this paper presents a \emph{Simplified VOS framework} (SimVOS) for joint feature extraction and matching. This basic idea allows SimVOS to learn dynamic target-aware features for more accurate VOS. Inspired by the recent successes on replacing \jimmy{hand-crafted} 
designs \cite{simtrack,ViT,ostrack} with general-purpose architectures in computer vision, we implement SimVOS with a ViT \cite{ViT} backbone and a mask prediction head. As can be seen in Fig.~\ref{archi_compare}, this new design removes the customized separate feature extraction and matching modules used in previous matching-based VOS approaches, thus facilitating the development of VOS in a more general and simpler system. 

Besides providing a simple yet effective VOS baseline, the other goal of our work is to bridge the gap between the VOS and large-scale self-supervised pretraining communities. Recently, significant progress \cite{mae,clip,moco,simclr} have been made in showing the superior performance of large-scale self-supervised models on some downstream tasks, including object classification \cite{mae}, detection \cite{plaindet} and tracking \cite{ostrack,simtrack}. However, existing VOS approaches often rely on task-specific customized modules, and their architectures are specifically designed with VOS-specific prior knowledge, which makes it difficult for these approaches to utilize standard large-scale self-supervised models for VOS. As far as we know, leveraging a large-scale self-supervised model to develop a general-purpose architecture for VOS has not been explored. Our work moves towards a simple yet effective VOS framework that could naturally benefit from large-scale self-supervised pretraining tasks.

\jimmy{Taking both memory and query frames as input to the vanilla ViT for mask prediction may cause a large computational cost (quadratic complexity).} 
To  achieve a better performance-speed trade-off, we propose a token refinement module to reduce the computational cost and improve the running speed of SimVOS. This variant can run $2\times$ faster than the SimVOS baseline, \abc{with a small reduction in VOS performance}. We conduct experiments on various popular VOS benchmarks and show that our SimVOS achieves state-of-the-art VOS performance.
In summary, this paper makes the following contributions:
\begin{compactitem}
  \item 
 We propose a \emph{Simplified VOS framework} (SimVOS), which removes the \jimmy{hand-crafted} feature extraction and matching modules in previous approaches \cite{STM,STCN}, to perform joint feature extraction and interaction via a single scalable transformer backbone. We also demonstrate that  large-scale self-supervised pre-trained models can provide significant benefits to the VOS task.
 \item
 We proposed a new token refinement module to achieve a better speed-accuracy trade-off for scalable video object segmentation.
  \item
  \jimmy{Our SimVOS achieves state-of-the-art performance on popular VOS benchmarks. Specifically, without applying any synthetic data pre-training, our variant SimVOS-B  sets new state-of-the-art performance on DAVIS-2017 (88.0\% $\mathcal{J}\&\mathcal{F}$), DAVIS-2016 (92.9\% $\mathcal{J}\&\mathcal{F}$) and YouTube-VOS 2019 (84.2\% $\mathcal{J}\&\mathcal{F}$).}
 \end{compactitem}

 \begin{figure*}
\begin{center}
   \includegraphics[width=1.0\linewidth]{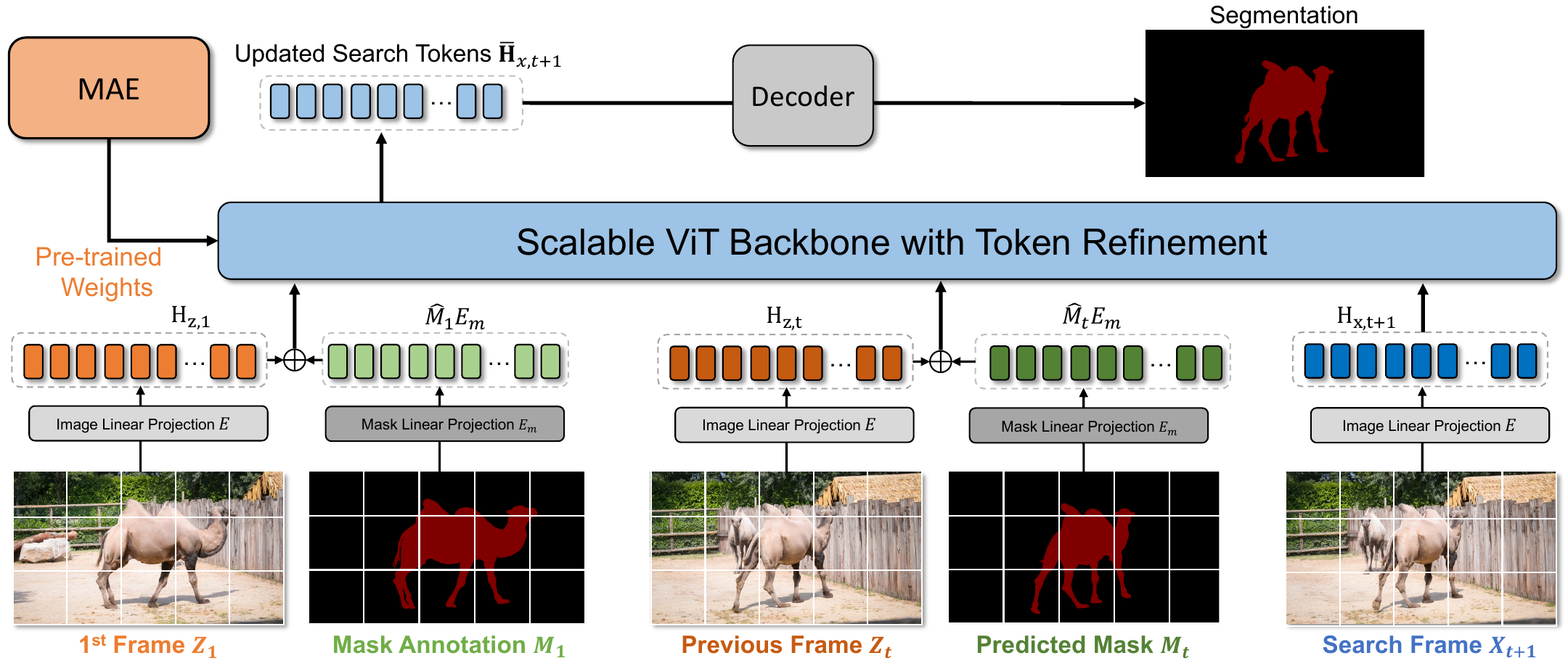} 
\end{center}
 \caption{The overall architecture of the proposed Simplified Video Object Segmentation (SimVOS) framework. Our SimVOS  consists of a scalable ViT backbone with token refinement for template and search token interaction, a decoder for segmentation mask prediction, and image/mask projection modules. }
\label{overall_framework}
\end{figure*}

\section{Related Work}
\noindent\textbf{Video Object Segmentation.} Traditional VOS methods follow the basic idea of online fine-tuning at test time to adapt 
to online tracked objects. Typical works \yty{include} OSVOS \cite{osvos}, OnAVIS \cite{OnaVOS}, MoNet \cite{monet}, MaskTrack \cite{MaskTrack} and PReMVOS \cite{PReMVOS}. However, the  time-consuming fine-tuning step limits their applicability to real-time applications, and meanwhile, the limited number of online training samples still degrades online fine-tuning. To improve the inference efficiency, OSMN \cite{OSMN} employs a meta neural network to guide mask prediction and uses a single forward pass to adapt the network to a specific test video. PML \cite{PML} formulates VOS as a pixel-wise retrieval task in the learned feature embedding space and uses a nearest-neighbor approach for real-time pixel-wise classification.

The typical matching-based approach STM \cite{STM} employs an offline-learned matching network, and treats past frame predictions as memory frames for the current frame matching. CFBI \cite{CFBI} and FEELVOS \cite{Feelvos} further improve the matching mechanism by leveraging foreground-background integration and local-global matching. AOT \cite{AOT} employs a long short-term transformer and an identification mechanism for the simultaneously multi-object association. STCN \cite{STCN} uses the L2 similarity to replace the dot product used in STM and establishes correspondences only on images to further improve the inference speed. To efficiently encode spatiotemporal cues, SSTVOS \cite{SSTVOS} uses a sparse spatiotemporal transformer. XMEM \cite{XMEM} further proposes an Atkinson-Shiffrin memory model to enable STM-like approaches to perform long-term VOS. 
Although  successes have been achieved by these matching-based approaches, their performance is still limited due to the separate feature extraction and interaction designs. In this work, we show that the above hand-crafted designs can be effectively replaced with a general-purpose transformer architecture, i.e., a vanilla ViT \cite{ViT} backbone with joint feature extraction and interaction, which can greatly benefit from existing large-scale pre-trained models (e.g., MAE \cite{mae}) and thus improve the state-of-the-art VOS performance \cite{davis16,davis17}.

\noindent\textbf{Large-scale self-supervised pre-training.} Large-scale self-supervised pre-training has achieved significant progress in recent years. Traditional approaches mainly focus on designing various pretext tasks for unsupervised representation learning, e.g., solving jigsaw puzzle \cite{jigsaw},  coloring images \cite{coloring} and predicting future frame \cite{future_prediction_1,future_prediction_2} or rotation angle \cite{rotation_pred}. Recent advances \cite{simclr,moco,membank,CMC} show that more discriminative unsupervised feature representations can be learned in a contrastive paradigm. However, these approaches may ignore modeling of local image structures, thus being limited in some fine-grained vision tasks, e.g., segmentation. The generative MAE \cite{mae} approach further improves on the contrastive learning-based methods by learning more fine-grained local structures, which are beneficial for localization or segmentation-based downstream tasks. In this work, our proposed SimVOS can directly apply the pre-trained models learned by existing self-supervised methods, which effectively bridges the gap between the VOS and the self-supervised learning communities. We also show that 
MAE 
can serve as a strong pre-trained model for the VOS task.
 

\section{Methodology}
In this section, we present our proposed \emph{Simplified VOS} (SimVOS) framework. An overview of SimVOS is shown in Fig.~\ref{overall_framework}. We firstly introduce the basic SimVOS baseline with joint feature extraction and interaction for accurate video object segmentation in Sec.~\ref{text:basic}. Then, in order to 
reduce the computational cost and improve the inference efficiency, multiple speed-up strategies are explored in Sec. \ref{speed_up}, including the usage of within-frame attention and a novel token refinement module for reducing the number of tokens used in the transformer backbone.



\subsection{Simplified Framework}
\label{text:basic}
As shown in Fig.~\ref{overall_framework}, our basic SimVOS mainly consists of a joint featrue extraction module and a mask prediction head. We use a vanilla ViT \cite{ViT} as the backbone of SimVOS, which is mainly because: 1) ViT naturally performs the joint feature extraction and interaction, which perfectly meets our design; and 2) a large amount of pre-trained ViT models can be directly leveraged in 
the VOS task, without needing time-consuming \abc{model-specific} synthetic video pre-training \jimmy{commonly used for previous VOS methods \cite{STCN,XMEM,STM}}.

In our SimVOS, the memory frames for online matching consist of 
the initial template frame $Z_{1} \in \mathbb{R}^{3\times H \times W}$ with ground truth mask $M_{1} \in \mathbb{R}^{H \times W}$ and the previous $t$-th frame ${Z}_{t} \in \mathbb{R}^{3\times H \times W}$ with the predicted mask $M_{t} \in \mathbb{R}^{H \times W}$. 
Given a current search frame  $X_{t+1}$, the goal of SimVOS is to accurately predict the mask of $X_{t+1}$ based on the memory frames. Instead of directly inputing the input frames to the ViT backbone for joint feature extraction and interaction, the input frames are first serialized into input sequences. Specifically, each input frame is reshaped into a sequence of flattened 2D patches with the size of $N \times 3P^{2}$ to obtain the reshaped memory sequences (i.e., $\hat{Z_{1}} \in \mathbb{R}^{N \times 3P^{2}}$ and $\hat{Z}_{t} \in \mathbb{R}^{N \times 3P^{2}}$) and the search sequence $\hat{X}_{t+1} \in \mathbb{R}^{N \times 3P^{2}}$, where $P\times P$ is the patch resolution and $N=HW/P^{2}$ is the number of patches. After applying the linear projection $\mathbf{E} \in \mathbb{R}^{3P^{2}\times C}$ to convert the 2D patches to 1D tokens with $C$ dimensions and adding the sinusoidal positional embedding \cite{ViT}, $\mathbf{P} \in \mathbb{R}^{N \times C}$, we get the memory embeddings, $\mathbf{H}_{z,1} \in \mathbb{R}^{N \times C}$ and $\mathbf{H}_{z,t} \in \mathbb{R}^{N \times C}$, and the search embeddings $\mathbf{H}_{x,t+1} \in \mathbb{R}^{N \times C}$.

 \begin{figure}
\begin{center}
   \includegraphics[width=0.9\linewidth]{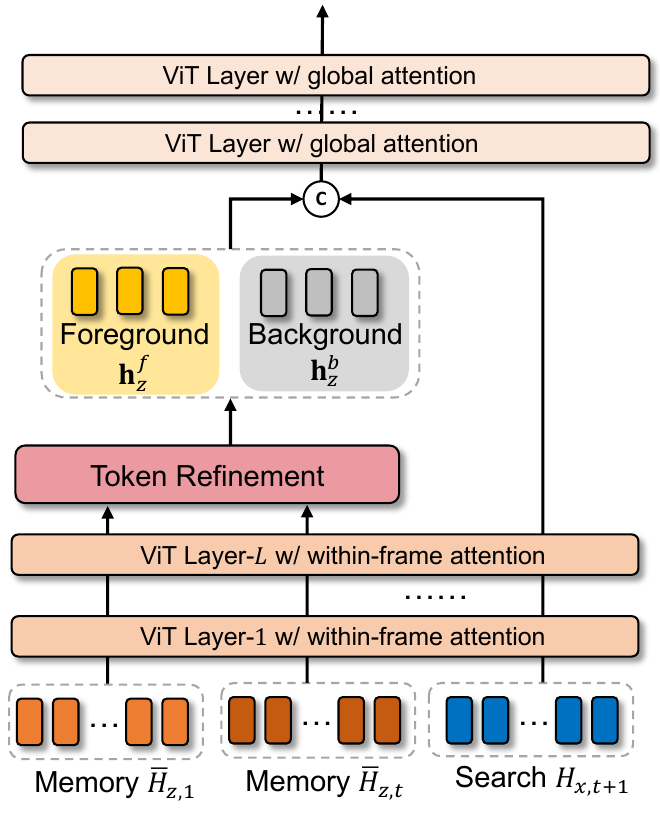} 
\end{center}
 \caption{Overall pipeline of a scalable ViT backbone with within-frame attention \abc{applied to the first $L$ layers}, and our token refinement module applied \abc{after} the $L$-th layer. 
 }
 \label{token_refine_overall}
\end{figure}

\noindent\textbf{Encoding mask annotation.} To encode the mask annotation in the joint feature extraction and interaction, we use a linear projection $\mathbf{E}_{m} \in \mathbb{R}^{P^2 \times C}$ to convert the 2D mask map to mask embeddings, which can be alternatively regarded as target-aware positional embeddings. Following the image sequence generation, the 2D mask maps \jimmy{$M_{1}$} 
and $M_{t}$ are firstly flattened into sequences, i.e.,  $\hat{M}_{1},\hat{M}_{t} \in \mathbb{R}^{N \times P^2}$. 
Next, the mask annotation are incorporated into the input memory embeddings,
\begin{align}\label{similarity}
\bar{\mathbf{H}}_{z,1} = \hat{M}_{1}\mathbf{E}_{m} + \mathbf{H}_{z,1},\\
\bar{\mathbf{H}}_{z,t} = \hat{M}_{t}\mathbf{E}_{m} + \mathbf{H}_{z,t}.
\end{align}
The obtained memory embeddings and search embeddings are concatenated together to form the input embeddings 
$\mathbf{H}^{0}=[\bar{\mathbf{H}}_{z,1};\bar{\mathbf{H}}_{z,t};\mathbf{H}_{x,t+1}]$ 
to the vanilla ViT for joint feature extraction and interaction.


\noindent\textbf{Joint feature extraction and matching.} Previous VOS approaches extract the features of memory and search frames firstly, and then employ a manually-designed matching layer to attend the memory features to the search features for the final mask prediction. 
In SimVOS, this feature extraction and matching step can be simultaneously implemented in a more elegant and general way via the multi-head self-attention used in the vanilla ViT.





\noindent\textbf{Mask Prediction.} The updated search embedding \jimmyy{$\bar{\mathbf{H}}_{x,t+1}$ output from the last layer of ViT} 
is further reshaped to a 2D feature map. Following the previous approach STM \cite{STM},  we use the same decoder that consists of several convolutional and deconvolutional layers for the final mask prediction. Since the decoder requires multi-resolution inputs, \yty{$\bar{\mathbf{H}}_{x,t+1}$} 
is firstly upsampled to $2\times$ and $4\times$ sizes via the deconvolution-based upsampling modules used in \cite{STM}. 

\subsection{Speed-up Strategies}
\label{speed_up}
Despite the favorable segmentation results achieved by the proposed basic SimVOS, the computational and memory complexity for multi-head attention on the long sequence input $\mathbf{H}^{0} \in \mathbb{R}^{3N \times C}$  can be very high when the frame resolution is large. To reduce the computational cost, \jimmy{we explore multiple speed-up strategies including within-frame attention and token refinement for foreground and background prototype generation. Fig. \ref{token_refine_overall} illustrates a scalable ViT backbone with our speed-up strategies.}

\noindent\textbf{Within-frame attention.} In the vanilla ViT, each query \yty{token} globally \yty{attends to all other tokens, }
thus leading to quadratic complexity.  However, it may be less necessary to perform the global \yty{token interaction} in the early layers since the shallow features mainly focus more on the local structures instead of catching the long-range dependency. Therefore, for each query \yty{token, we restrict token attendance to only those within the same frame, and refer to this as \emph{within-frame attention}.} 
By applying the within-frame attention, the complexity of the computation in a specific layer reduces from $\mathcal{O}(9N^{2})$ to $\mathcal{O}(N^{2})$. In practice, \yty{within-frame }
attention is used in the first $L$ layers of ViT. 


\noindent\textbf{Token refinement module.} The within-frame attention reduces the overall complexity in the first  $L$ layers of ViT. However, the global self-attention used in the rest of the layers in ViT still causes the large quadratic complexity. \yty{Continuing to perform within-frame attention for deep layers}
alleviates this issue but also causes significant performance degradation, which is illustrated in Table \ref{speed_accuracy_table}. To address the aforementioned issues, we propose a token refinement module to further reduce the number of tokens in the memory embeddings, thus leading to a significant reduction of computational cost in the global self-attention layers.

\begin{figure}
\begin{center}
   \includegraphics[width=0.9\linewidth]{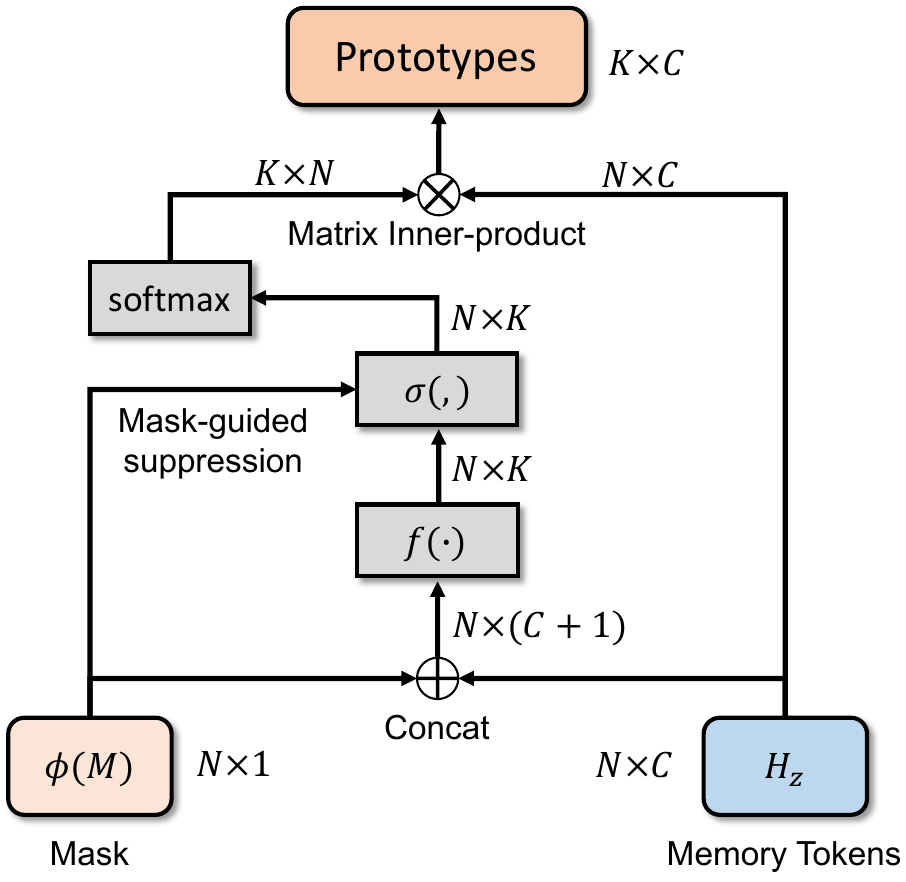} 
\end{center}
 \caption{Detailed pipeline of the token refinement module, which  effectively generates a small number of foreground and background prototypes guided by mask distribution.}
 \label{proto_generation}
\end{figure}

Recent advances \cite{patch_to_cluster,tokenlearner,tore} in token reduction for efficient image classification show that the tokens can be effectively clustered by using a learnable convolutional module. In our work, given a memory embedding $\mathbf{H}_{z}  \in \mathbb{R}^{N \times C}$ \jimmyy{extracted from the first $L$ layers of ViT using within-frame attention.}
our goal is to cluster or segment $\mathbf{H}_{z}$ into several foreground and background prototypes, \jimmy{in order to reduce the overall complexity in the following global self-attention layers}. This can be achieved by proposing a learnable token refinement module for prototype generation guided by the segmentation mask $M$.
Specifically, the generation of foreground prototypes can be formulated as:
\begin{align}\label{prot_generation}
W=f([\mathbf{H}_{z},\phi(M)])  \in \mathbb{R}^{N \times K},\\
\label{prot_generation2}
\hat{W} = \text{softmax}(\sigma(W, \phi(M)))  \in \mathbb{R}^{N \times K},
\end{align}
where \yty{$K$ is the number of generated prototypes. }$[,]$ is the concatenation operation at the channel dimension, \jimmy{the softmax function is applied over each column of the input 2D matrix $\in \mathbb{R}^{N \times K}$}, and $\phi()$ denotes a downsampling operation to reshape  the mask $M$ in order to meet the same spatial size with $\mathbf{H}_{z}$, i.e., $\phi(M) \in \mathbb{R}^{N \times 1}$. $f(\cdot)$ is a clustering function, which is implemented as a neural network module that consists of several convolutional layers and a fully-connected layer, \jimmy{(see supplementary for details)}. Inspired by \cite{patch_to_cluster,tokenlearner}, the convolutional layers are firstly employed to map the high-dimensional input features to lower feature dimension, and the fully-connected layer predicts a prototype assignment matrix $\hat{W}$ in order to map the original features to $K$ latent prototypes. We also use a post-processing function $\sigma(\cdot) $ to suppress weights at non-target locations in $W$, which is achieved by setting the corresponding rows in $W$ to negative infinity, such that these elements can be suppressed after applying the softmax function.
Finally, the generated prototypes are generated as:
\begin{align}\label{similarity}
\mathbf{h}_{z} = \hat{W}^{T} \mathbf{H}_{z} \in \mathbb{R}^{K \times C}.
\end{align}
For the background prototype generation, we simply replace the mask $M$ used in (\ref{prot_generation}) and (\ref{prot_generation2}) with the reverse mask i.e., $1-M$. \yty{For clarification, we denote the foreground and background prototypes as $\mathbf{h}^f_{z}\in \mathbb{R}^{K_{f} \times C}$ and $\mathbf{h}^b_{z}\in \mathbb{R}^{K_{b} \times C}$ respectively, where $K_f$ and $K_b$ indicate their corresponding number of generated prototypes.} 
\jimmy{As shown in Fig.~\ref{token_refine_overall}, we then \yty{feed the concatenation of $\mathbf{h}^f_{z}$, $\mathbf{h}^b_{z}$ and the search tokens $\mathbf{H}_{x,t+1}$} to the remaining layers of ViT for global self-attention calculation.}

\jimmy{In Table \ref{num_prototype}}, we explore multiple settings of these two hyper-parameters $K_{f}$ and $K_{b}$, which can be set to relatively small numbers without degrading  performance much. Since $K_{f}+K_{b} \ll N$, the overall complexity is further reduced in the global self-attention layers based on the proposed token refinement module. For example, given a memory frame with the size of $480\times960\time3$, there are $1800$ tokens in total when $P=16$. Based on our method, only 512 prototypes are generated ($K_{f}=K_{b}=256$), which is about 3.5 times less than the original variant and achieves a better speed-accuracy trade-off.

We show the overall pipeline of our token refinement module in Fig. \ref{proto_generation} and \jimmy{further visualize the  assignment probability matrix $\hat{W}$ for both foreground and background prototype generation in Fig. \ref{simple_demo}}. Interestingly, we find that the token refinement (TR) module aims to aggregate boundary features for prototype generation. This observation provides a potential explanation on which kinds of spatial features are more beneficial for online matching in video object segmentation. It shows that the boundary structures provide more useful cues for accurate online segmentation.





 \vspace{-0.1cm}
\subsection{Training and Inference}
\noindent\textbf{Training on video datasets.} Previous approaches \cite{STM,STCN,XMEM,AOT} \jimmyy{commonly adopt two-stage or three-stage training including synthetic data pre-training using static image datasets \cite{coco,salient_image,pascal,Semantic_images,cssd}, BL30K \cite{bl30k} pre-training and main training on video datasets.} 
In this work, we observe that the proposed SimVOS can be well learned by only using the single stage of main training on video datasets (e.g., DAVIS2017 \cite{davis17} and YouTube-VOS 2019 \cite{youtubevos}), which further simplifies the training process. 
Specifically, we randomly sample two frames in a video clip with a predefined maximum frame gap (i.e., 10), in order to construct the template and search frames during the training. Following the convention \cite{STM,STCN}, the same bootstrapped cross entropy loss is used for supervision.

\noindent\textbf{Online inference.} During the online inference, we use the first frame and the predicted previous frame as the memory frames. The overall pipeline is shown in Fig. \ref{overall_framework}. There are no additional online adaptation or fine-tuning steps used.



\begin{table}[t]
  \newcommand{\tabincell}[2]
    \centering
    \begin{tabular}{cccc}
    \Xhline{\arrayrulewidth}
  \multicolumn{1}{c}{Pre-trained Method}  & \multicolumn{1}{c}{$\mathcal{J}\&\mathcal{F}\uparrow$} & \multicolumn{1}{c}{$\mathcal{J}\uparrow$} & \multicolumn{1}{c}{$\mathcal{F}\uparrow$} \cr
       \Xhline{\arrayrulewidth}
       Random & 68.8 & 66.3 & 71.3 \cr
       ImageNet1K \cite{deit} & 81.3 & 78.8 & 83.8 \cr
       ImageNet22K \cite{imagenet21k_learn} & 83.5 & 80.5 & 86.6 \cr
       MoCo-V3 \cite{moco} & 81.5 & 79.0 & 83.9 \cr
       MAE \cite{mae} & \textbf{88.0} & \textbf{85.0} & \textbf{91.0} \cr
   
   \Xhline{\arrayrulewidth}  
   \end{tabular} 
  \vspace{-0.1cm}
   \centering
  \caption{The ablation study on the DAVIS-2017 val set using various pre-trained models for SimVOS with ViT-Base backbone.}  
  \vspace{-0.2cm}
  \label{network_init}
\end{table}

\newcommand{\cmark}{\ding{51}}%
\begin{table}[t]
  \newcommand{\tabincell}[2]
    \centering
  \footnotesize
    \begin{tabular}{c|cc|ccc|c}
    \Xhline{\arrayrulewidth}
 \multicolumn{1}{c|}{Backbone} & \multicolumn{1}{c}{$L$} & \multicolumn{1}{c|}{TR} & \multicolumn{1}{c}{$\mathcal{J}\&\mathcal{F}\uparrow$} & \multicolumn{1}{c}{$\mathcal{J}\uparrow$} & \multicolumn{1}{c|}{$\mathcal{F}\uparrow$}  & \multicolumn{1}{c}{FPS} \cr
       \Xhline{\arrayrulewidth}
   ViT-Base & 0 & & 88.0\% & 85.0\% & 91.0\% & 4.9 \cr
   ViT-Large & 0 && 88.5\% & 85.4\%& 91.5\%& 2.0 \cr
    \Xhline{\arrayrulewidth}
    ViT-Base & 2 & & 87.4\% & 84.4\% & 90.4\% & 5.7 \cr
    ViT-Base & 2 &\cmark & 86.0\% & 83.2\% & 88.9\% & 9.4 \cr
    ViT-Base & 4 & & 86.9\% & 84.0\% & 89.8\% & 6.5 \cr
    ViT-Base & 4 & \cmark& 87.1\% & 84.1\% &  90.1\% & 9.9 \cr
     ViT-Base & 6 & & 86.8\% & 83.7\% & 89.9\% & 7.6 \cr
     ViT-Base & 6 & \cmark& 86.5\% & 83.4\% & 89.5\% & 10.7 \cr
     ViT-Base & 8 & & 86.5\% &83.6\% & 89.4\% & 8.8 \cr
     ViT-Base & 8 & \cmark& 86.0\% & 83.0\% & 89.1\% & 11.4 \cr
   \Xhline{\arrayrulewidth}  
   \end{tabular} 
   \centering
  \caption{The performance of SimVOS variants on  the DAVIS-2017 validation set. $L$ denotes the number of layers using within-frame attention. TR indicates the usage of the token refinement module, where the default numbers of generated foreground/background prototypes are 384/384.
  }  
  \vspace{-0.1cm}
  \label{speed_accuracy_table}
\end{table}

\section{Implementation Details}
\noindent\textbf{Evaluation metric.} We use the official evaluation metrics, $\mathcal{J}$ and $\mathcal{F}$ scores, to evaluate our method. Note $\mathcal{J}$ is calculated as the average IoU between the prediction and ground-truth masks.  $\mathcal{F}$ measures the boundary similarity measure between the prediction and ground-truth. The $\mathcal{J}\&\mathcal{F}$ score is the average of the above two metrics. 

\noindent\textbf{Training and evaluation.} The proposed SimVOS is evaluated on multiple VOS benchmarks: DAVIS-2016 \cite{davis16}, DAVIS-2017 \cite{davis17} and YouTube-VOS 2019 \cite{youtubevos}. For a fair comparison with previous works \cite{STM,AOT}, we train our method on the training set of YouTube-VOS 2019 for YouTube-VOS evaluation. For DAVIS evaluation, we train SimVOS on both DAVIS-2017 and YouTube-VOS 2019. In the evaluation stage, we use the default 480P 24 FPS videos for DAVIS and 6 FPS videos for YouTube-VOS 2019 on an NVIDIA A100 GPU.

We use randomly cropped $384 \times 384$ patches from video frames for training.  We use the mini-batch size of 32 with a learning rate of $2\mathrm{e}{-5}$. The number of training iterations is set to $210,000$ and the learning rate is decayed to $2\mathrm{e}{-4}$ at half the iterations. The predefined maximum frame gap is set to a fixed value, i.e., 10, without using the curriculum learning schedule as described in \cite{STM,STCN,AOT} for simplicity. More training details can be found in the supplementary.

\section{Experiments}
In this section, we conduct ablation studies, state-of-the-art comparison and qualitative visualization to demonstrate the effectiveness of our proposed SimVOS.

\begin{table}[t]
  \newcommand{\tabincell}[2]
    \centering
    \begin{tabular}{cc|ccc|c}
    \Xhline{\arrayrulewidth}
  \multicolumn{1}{c}{$K_{f}$} & \multicolumn{1}{c|}{$K_{b}$} & \multicolumn{1}{c}{$\mathcal{J}\&\mathcal{F}\uparrow$} & \multicolumn{1}{c}{$\mathcal{J}\uparrow$} & \multicolumn{1}{c|}{$\mathcal{F}\uparrow$}  & \multicolumn{1}{c}{FPS} \cr
       \Xhline{\arrayrulewidth}
       \textbf{384} & \textbf{384} & 87.1\% & 84.1\% &  90.1\% & 9.9 \cr
       256 & 256 & 86.7\% & 83.6\% &  89.8\% & 11.4 \cr
       128 & 128 & 85.3\% & 82.1\% &  88.4\% & 13.5 \cr
       \Xhline{\arrayrulewidth}
       512 & 256 & 86.0\% & 83.0\% &  89.1\% & 9.9 \cr
       256 & 512 & 86.6\% & 83.7\% &  89.5\% & 9.9 \cr
   
   \Xhline{\arrayrulewidth}  
   \end{tabular} 
  \vspace{-0.1cm}
   \centering
  \caption{The ablation study on the number of generated foreground ($K_{f}$) and background ($K_{b}$) prototypes used in the TR module of our SimVOS, which employs the ViT-Base backbone and uses the first L=4 layers for within-frame attention. \jimmyy{The default prototype number used in SimVOS is shown in bold.}}
  \label{num_prototype}
\end{table}

\begin{table}[t]
  \newcommand{\tabincell}[2]
    \centering
    \begin{tabular}{cccc}
    \Xhline{\arrayrulewidth}
  \multicolumn{1}{c}{Variant}  & \multicolumn{1}{c}{$\mathcal{J}\&\mathcal{F}\uparrow$} & \multicolumn{1}{c}{$\mathcal{J}\uparrow$} & \multicolumn{1}{c}{$\mathcal{F}\uparrow$} \cr
       \Xhline{\arrayrulewidth}
       w/o $\sigma(,)$ & 84.9 & 81.9 & 88.0  \cr
       w/ $\sigma(,)$ & 87.1 & 84.1 & 90.1 \cr
   
   \Xhline{\arrayrulewidth}  
   \end{tabular}
   \centering
  \caption{The ablation study on the DAVIS-2017 val set w/ and w/o using the post-processing function $\sigma(,)$ in Eq. (4).}  
  \label{abl_sigma}
\end{table}

 \begin{figure}
\begin{center}
   \includegraphics[width=0.9\linewidth]{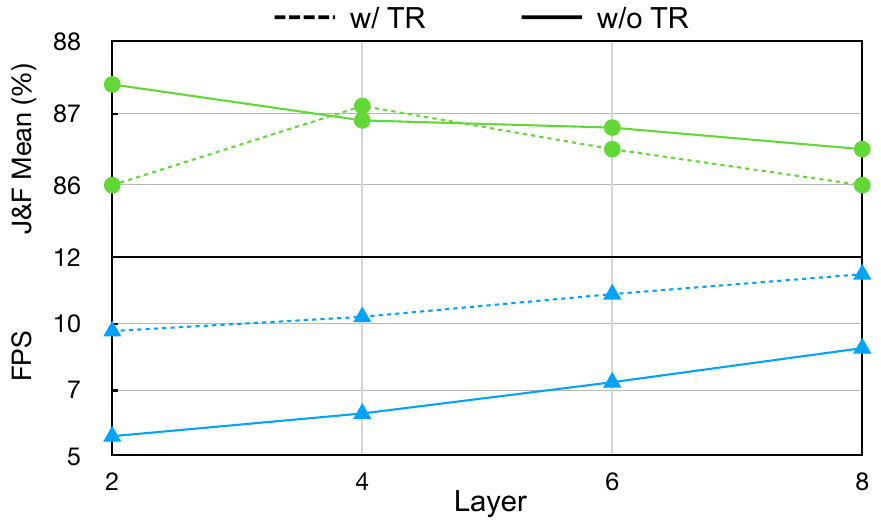} 
\end{center}
\vspace{-0.55cm}
 \caption{\jimmyy{The plot of speed and $\mathcal{J}\&\mathcal{F}$ score versus layer index $L$ for within-frame atttention. The dashed and solid lines represent the variant w/ and w/o applying the TR module, respectively.}}
 \label{speed_accuracy_plot}
\end{figure}

\subsection{Ablation Study}
\noindent\textbf{Large-scale pre-trained models.} We test several popular large-scale pre-trained models for initializing the ViT-Base backbone used in our SimVOS. As can be seen in Table \ref{network_init}, the generative MAE \cite{mae} model is the optimal choice compared with the other pre-trained models, including the supervised ImageNet1K \cite{deit}, ImageNet22K \cite{imagenet21k_learn} and the contrastive learning-based approach MoCoV3 \cite{moco}. This is mainly because fine-grained local structures are learned in MAE, which is more beneficial for the pixel-wise VOS task. Training with random initialization severely degrades the performance, which indicates that the number of training videos in DAVIS and YouTube-VOS is not sufficient enough for learning a robust VOS model. Based on these observations, we use the MAE pre-trained model as the default initialization for our SimVOS.

\begin{table}[t]
  \newcommand{\tabincell}[2]
    \centering
    \begin{tabular}{cccccc}
    \Xhline{\arrayrulewidth}
    \multirow{2}{*}{Method}&  
  \multirow{2}{*}{OL}& \multirow{2}{*}{S}&
  \multirow{2}{*}{$\mathcal{J}\&\mathcal{F}\uparrow$} &
  \multirow{2}{*}{$\mathcal{J}\uparrow$} &
  \multirow{2}{*}{$\mathcal{F}\uparrow$}\cr
  \cr
       \Xhline{\arrayrulewidth}
   OSVOS-S \cite{OSMN} & \cmark & \cmark & 68.0 & 64.7 & 71.3\cr
   OSVOS \cite{osvos} & \cmark &  & 60.3 & 56.6 & 63.9 \cr
   OnAVOS \cite{OnaVOS} & \cmark &  & 65.4 & 61.6 & 69.1\cr
   CINM \cite{cinm} & \cmark &  & 70.6 & 67.2 & 74.0\cr
   AFB-URR \cite{afb_urr} &  & \cmark & 76.9 & 74.4 & 79.3\cr
   STM \cite{STM} & & \cmark & 81.8 & 79.2 & 84.3\cr
   RMNet \cite{RMNet} &  & \cmark & 83.5 & 81.0 & 86.0\cr
   HMMN \cite{HMMN} &  & \cmark & 84.7 & 81.9 & 87.5\cr   
   MiVOS$^{*}$ \cite{bl30k} &  & \cmark & 84.5 & 81.7 & 87.4\cr    
   STCN \cite{STCN} &  & \cmark & 85.4 & 82.2 & 88.6\cr   
   STCN$^{*}$ \cite{STCN} &  & \cmark & 85.3 & 82.0 & 88.6\cr    
   AOT \cite{AOT} & & \cmark & 84.9 & 82.3 & 87.5\cr       
   XMEM \cite{XMEM} &  & \cmark & 86.2 & 82.9 & 89.5\cr
   XMEM$^{*}$ \cite{XMEM} &  & \cmark & 87.7 & 84.0 & \underline{91.4}\cr
   \Xhline{\arrayrulewidth}
   CFBI \cite{CFBI} &  &  & 81.9 & 79.3 & 84.5\cr
    JOINT \cite{joint} &  &  & 83.5 & 80.8 & 86.2\cr      
    SSTVOS \cite{SSTVOS} & & & 82.5 & 79.9 & 85.1 \cr
   FAVOS \cite{favos} & & &58.2 & 54.6 & 61.8 \cr 
   STCN$^{-}$ \cite{STCN} &&&  82.5 & 79.3 & 85.7 \cr
   XMEM$^{-}$ \cite{XMEM} &  &  & 84.5 & 81.4 & 87.6\cr
  \jimmyy{\textbf{SimVOS-BS}} &  &  & 87.1 & 84.1 &  90.1 \cr
   \textbf{SimVOS-B} &  &  & \underline{88.0} & \underline{85.0} & 91.0\cr
   \textbf{SimVOS-L} &  &  & \textbf{88.5} & \textbf{85.4} & \textbf{91.5}\cr
   \Xhline{\arrayrulewidth}  
   \end{tabular} 
   \centering
  \caption{ Comparisons with previous approaches on the DAVIS-2017 validation set. OL and S represent the online learning and synthetic data pre-training. $*$ denotes the BL30K \cite{bl30k} pre-training. $-$ means to remove synthetic data pre-training.} 
  \vspace{-0.5cm}
  \label{davis17_compare}
\end{table}

\begin{table}[t]
  \newcommand{\tabincell}[2]
    \centering
    \begin{tabular}{cccccc}
    \Xhline{\arrayrulewidth}
    \multirow{2}{*}{Method}&  
  \multirow{2}{*}{OL}& \multirow{2}{*}{S}&
  \multirow{2}{*}{$\mathcal{J}\&\mathcal{F}\uparrow$} &
  \multirow{2}{*}{$\mathcal{J}\uparrow$} &
  \multirow{2}{*}{$\mathcal{F}\uparrow$}\cr
  \cr
       \Xhline{\arrayrulewidth}
   OSVOS \cite{osvos} & \cmark &  & 80.2 & 79.8 & 80.6 \cr    
   OnAVOS \cite{OnaVOS} & \cmark &  & 85.7 & - & - \cr
   CINM \cite{cinm} & \cmark &  & 84.2 & - & -\cr
   STM \cite{STM} & & \cmark & 89.3 & 88.7 & 89.9\cr   
   RMNet \cite{RMNet} &  & \cmark & 88.8 & 88.9 & 88.7\cr    
   HMMN \cite{HMMN} &  & \cmark & 90.8 & 89.6 & 92.0\cr      
   MiVOS$^{*}$ \cite{bl30k} &  & \cmark & 91.0 & 89.6 & 92.4\cr    	   
   STCN \cite{STCN} &  & \cmark & 91.6 & 90.8 & 92.5\cr        
   STCN$^{*}$ \cite{STCN} &  & \cmark & 91.7 & 90.4 & 93.0\cr      
   AOT \cite{AOT} & & \cmark & 91.1 & 90.1 & 92.1\cr         
   XMEM \cite{XMEM} &  & \cmark & 91.5 & 90.4 & 92.7\cr      
   XMEM$^{*}$ \cite{XMEM} &  & \cmark & 92.0 & 90.7 & 93.2\cr	  
   \Xhline{\arrayrulewidth}
   FAVOS \cite{favos} & & &-& 82.4 & 79.5 \cr 
   CFBI \cite{CFBI} &  &  & 89.4 & 88.3 & 90.5\cr   
   XMEM$^{-}$ \cite{XMEM} &  &  & 90.8 & 89.6 & 91.9\cr	  
  \jimmyy{\textbf{SimVOS-BS}} &  &  & 91.5 & 89.9 &  93.1 \cr 
   \textbf{SimVOS-B} &  &  & \underline{92.9} & \underline{91.3} & \underline{94.4}\cr		
   \textbf{SimVOS-L} &  &  & \textbf{93.6} & \textbf{92.0} & \textbf{95.3}\cr	
   \Xhline{\arrayrulewidth}  
   \end{tabular} 
   \centering
  \caption{ Comparisons with previous approaches on the DAVIS-16 validation set. OL and S represent the online learning and synthetic data pre-training. $*$ denotes the BL30K \cite{bl30k} pretraining. $-$ means without applying synthetic data pre-training.} 
   \vspace{-0.15cm}
  \label{davis16_compare}
\end{table}

\noindent\textbf{Within-frame attention.}  The within-frame attention (i.e., $L=[2,4,6,8]$) can reduce the overall computational cost and improve the inference speed. As shown in Table \ref{speed_accuracy_table} and Fig. \ref{speed_accuracy_plot}, when the number of within-frame attention layers $L$ is increased, the variants are more efficient but also suffer from some performance degradation, due to the insufficient interaction between the memory and search tokens. Considering the performance-speed trade-off, we use $L=4$ as our \yty{default setting} for further token refinement.

\noindent\textbf{Token refinement.} The token refinement (TR) module can further reduce the overall complexity in the global multi-head self-attention layers of SimVOS. There are several observations in Table \ref{speed_accuracy_table} and Fig.~\ref{speed_accuracy_plot}: 1) the TR module applied in the early layer (e.g., $L=2$) may cause insufficient memory token encoding, thus leading to large performance drop; 2) When $L\geq 4$, the TR module improves the inference speed, and achieves comparable results with the baseline.

\noindent\textbf{Number of prototypes.} The number of generated foreground or background prototypes may affect the overall performance of SimVOS. We conduct this ablation study in Table \ref{num_prototype}. As can be seen, severely decreasing the prototype number from 384 to 128 causes a relatively large drop of 1.8\% $\mathcal{J}\&\mathcal{F}$. This shows that the TR module needs enough prototypes (e.g., 384) to represent large foreground or background regions in a video frame. 

\noindent\textbf{Ratio of foreground and background prototypes.} We fix the total number (i.e., 768) of foreground and background prototypes, and test different ratios (i.e., 1:1, 2:1, and 1:2) of these two types of prototypes in Table \ref{num_prototype}. We find that the variant with balanced foreground and background prototypes achieves the best performance, which is because the foreground and background prototypes are all essential for accurate VOS in future frames of a test video.  

\noindent\textbf{Impact of $\sigma(,)$}. In Table \ref{abl_sigma}, we study the impact of with (w/) or without (w/o) the usage of mask map for foreground or background prototype generation. For the variant w/o using the mask map, we remove the post-processing function $\sigma(,)$ in Eq. 4 and directly use $W$ for prototype generation.
As we can see, w/o the usage of mask map, the generated prototypes are low quality, which degrades the performance with a margin of xx in terms of the $\mathcal{J}\&\mathcal{F}$ metric. This is mainly because this arbitrary generation may fix both foreground and background regions, thus generating ambiguous prototypes and causing more tracking failures.

 \begin{figure*}
\begin{center}
   \includegraphics[width=0.95\linewidth]{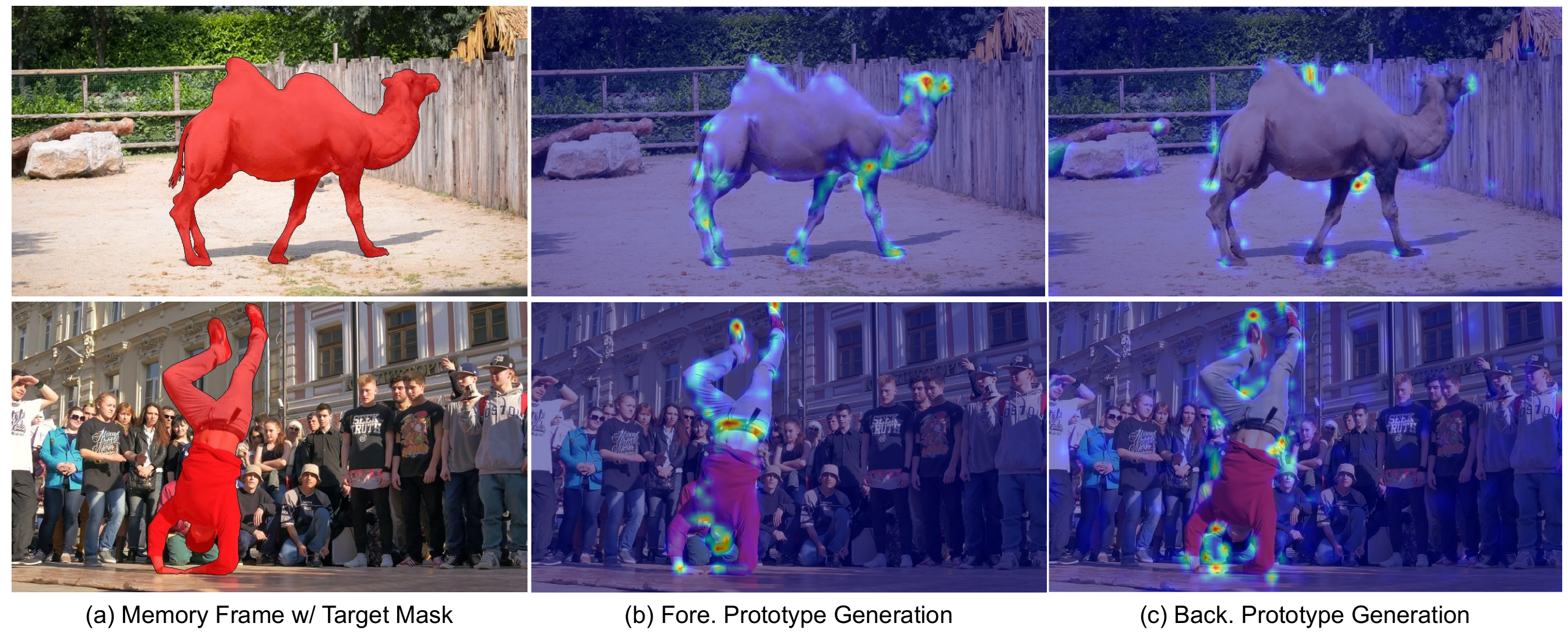} 
\end{center}
 \caption{\jimmy{Given the (a) memory frame w/ target mask, the visualization of assignment matrix $\hat{W} \in \mathbb{R}^{N \times K}$ (Eq. \ref{prot_generation2}) for both (b) foreground and (c) background prototype generation. $\hat{W} \in $ is averaged over each row and then upsampled to the image size for visualization. The TR module tends to aggregate boundary features to generate prototypes for accurate online VOS. More visualization is shown in supplementary.}}
 \label{simple_demo}
\end{figure*}

\subsection{State-of-the-art Comparison}
In this section, we compare multiple variants of our SimVOS with state-of-the-art VOS approaches. Specifically, the variants employ the VIT-Base or ViT-Large as the backbone, 
and do not use within-frame attention or the TR module, 
which are respectively denoted as \textbf{SimVOS-B} and \textbf{SimVOS-L}. We also include a full-version variant that employs the ViT-Base backbone and \jimmyy{all the speed-up strategies} including the within-frame attention and the TR module, denoted as \jimmyy{\textbf{SimVOS-BS}}.

\noindent\textbf{DAVIS-2017.} DAVIS-2017 \cite{davis17} is a typical VOS benchmark which has been widely for state-of-the-art comparisons in the VOS community. This dataset consists of 150 sequences and 376 annotated objects in total.  The validation set of DAVIS-2017 contains 30 challenging videos and uses a multi-object setting, where multiple annotated objects in the initial video frame are required to track and segment in the following frames. \jimmyy{The test set contains more challenging 30 videos for evaluation. }

\begin{table}[t]
  \newcommand{\tabincell}[2]
    \centering
    \begin{tabular}{ccccc}
    \Xhline{\arrayrulewidth}
    \multirow{2}{*}{Method} & \multirow{2}{*}{S}&
  \multirow{2}{*}{$\mathcal{J}\&\mathcal{F}\uparrow$} &
  \multirow{2}{*}{$\mathcal{J}\uparrow$} &
  \multirow{2}{*}{$\mathcal{F}\uparrow$}\cr
  \cr
       \Xhline{\arrayrulewidth}
   RMNet \cite{RMNet} & \cmark & 75.0 & 71.9 & 78.1\cr
   STCN \cite{STCN}   & \cmark & 76.1 & 73.1 & 80.0\cr      
   STCN$^{*}$ \cite{STCN}   & \cmark & 79.9 & 76.3 & 83.5\cr 
   AOT  \cite{AOT}   & \cmark & 79.6 & 75.9 & 83.3\cr       
   MiVOS$^{*}$ \cite{bl30k}  & \cmark & 78.6 & 74.9 & 82.2\cr     
   XMEM \cite{XMEM}   & \cmark & 81.0 & 77.4 & 84.5 \cr	     
   XMEM$^{*}$ \cite{XMEM}   & \cmark & 81.2 & 77.6 & 84.7 \cr	   
   \Xhline{\arrayrulewidth}
   CFBI \cite{CFBI}   &  & 75.0 & 71.4 & 78.7\cr     
   CFBI$+$ \cite{cfbi_plus}  & & 78.0 & 74.4 & 81.6 \cr   
   XMEM$^{-}$ \cite{XMEM}   &  & 79.8 & \underline{76.3} & 83.4 \cr	    
   \textbf{SimVOS-BS}   &  & 79.3 & 75.1 &  83.6 \cr 
   \textbf{SimVOS-B}   &  & \underline{80.4} &76.1 & \underline{84.6}\cr		
   \textbf{SimVOS-L}   &  & \textbf{82.3} & \textbf{78.7} & \textbf{85.8}\cr	
   \Xhline{\arrayrulewidth}  
   \end{tabular} 
   \centering
  \caption{ \jimmyy{Comparisons with previous approaches on the DAVIS-2017 test-dev. S indicates the usage of synthetic data pre-training. $*$ denotes the BL30K \cite{bl30k} pre-training. $-$ means without applying synthetic data pre-training. We use 480p videos for evaluation.}}
  \label{davis17_testdev_compare}
\end{table}

\begin{table}[t]
  \newcommand{\tabincell}[2]
    \centering
    \begin{tabular}{ccccc}
    \Xhline{\arrayrulewidth}
    \multirow{2}{*}{Method} & \multirow{2}{*}{S}&
  \multirow{2}{*}{$\mathcal{J}\&\mathcal{F}\uparrow$} &
  \multirow{2}{*}{$\mathcal{J}_{seen}\uparrow$} &
  \multirow{2}{*}{$\mathcal{J}_{unseen}\uparrow$}\cr
  \cr
       \Xhline{\arrayrulewidth}
   MiVOS$^{*}$ \cite{bl30k}  & \cmark & 82.4 & 80.6 & 78.1\cr  
   HMMN \cite{HMMN}   & \cmark & 82.5 & 81.7 & 77.3\cr      
   STCN \cite{STCN}   & \cmark & 82.7 & 81.1 & 78.2\cr      
   STCN$^{*}$ \cite{STCN}   & \cmark & 84.2 & 82.6 & 79.4\cr      
   SwinB-AOT \cite{AOT}   & \cmark & 84.5 & 84.0 & 78.4\cr  
   XMEM \cite{XMEM}   & \cmark & 85.5 & 84.3 & 80.3 \cr	  
   XMEM$^{*}$ \cite{XMEM}   & \cmark & 85.8 & 84.8 & 80.3 \cr	    
   \Xhline{\arrayrulewidth}
   CFBI \cite{CFBI}   &  & 81.0 & 80.6 & 75.2\cr   
   CFBI$+$ \cite{cfbi_plus}  & & 82.6 & 81.7 & 77.1 \cr
   JOINT \cite{joint} && \underline{82.8} & 80.8 & \underline{79.0} \cr
   SSTVOS \cite{SSTVOS}  & & 81.8 & 80.9 & 76.7 \cr
   XMEM$^{-}$ \cite{XMEM}   &  & \textbf{84.2} & \textbf{83.8} & 78.1 \cr	  
   \textbf{SimVOS-B}   &  & \textbf{84.2} & \underline{83.1} & \textbf{79.1}\cr		
   \Xhline{\arrayrulewidth}  
   \end{tabular} 
   \centering
  \caption{ Comparisons with previous approaches on the YouTube-VOS 2019 validation set. S indicates the usage of synthetic data pre-training. $*$ denotes the BL30K \cite{bl30k} pretraining. $-$ means without applying synthetic data pre-training.} 
  \label{ytvos_compare}
\end{table}

The comparison between our SimVOS and previous approaches on the DAVIS-2017 validation set is shown in Table \ref{davis17_compare}. Without applying online learning and synthetic data pre-training, our SimVOS-B and SimVOS-L  set new state-of-the-art $\mathcal{J}\&\mathcal{F}$ scores, i.e., 88.0\% and 88.5\%, which are even better than XMEM$^{*}$ that applies  both BL30K and  synthetic data pre-training. 
\jimmyy{The results on DAVIS-2017 test-dev is shown in Table \ref{davis17_testdev_compare}, our SimVOS-BS achieves comparable results to XMEM$^{-}$ that employs more memory frames. The SimVOS-L variant achieves 82.3\% $\mathcal{J}\&\mathcal{F}$ score, which is the leading performance on this dataset by using 480P videos for evaluation.}
We believe the strong performance of SimVOS can be attributed to two main aspects: 1) the generative MAE initialization, and 2) the  ViT backbone is suitable for memory and search interaction.

\noindent\textbf{DAVIS-2016.} DAVIS-2016 is a subset of DAVIS-2017 and it follows a single-object setting. For completeness, we also compare SimVOS with state-of-the-art approaches on the DAVIS-2016 validation set, which is illustrated in Table \ref{davis16_compare}. The $\mathcal{F}$ scores achieved by our SimVOS-F, SimVOS-B and SimVOS-L are  93.1\%, 94.4\% and 95.3\%, respectively. These results are significantly better than the previous SOTA approaches under the same training settings.



\noindent\textbf{YouTube-VOS 2019.} YouTube-VOS 2019 \cite{youtubevos} is a large-scale VOS benchmark that consists of 507 validation videos for evaluation. In Table \ref{ytvos_compare}, we show that our SimVOS performs favorably against state-of-the-art approaches on the YouTube-VOS 2019 validation set. Note that our SimVOS-B is evaluated at the default 6 FPS videos, but still achieves comparable performance with XMEM$^{-}$ that uses all the frames for evaluation. Moreover, the $\mathcal{J}_{unseen}$ score of SimVOS-B is 79.1\%, outperforming the others under the same setting (i.e., w/o synthetic data pre-training), which shows our method can generalize well to unseen objects during the testing.
\jimmy{More  results on YouTube-VOS 2019 and qualitative visualization are included in the supplementary.}



\section{Conclusion}
In this work, we present a scalable video object segmentation approach with simplified frameworks, called SimVOS. Our SimVOS removes hand-crafted designs (e.g., the matching layer) used in previous approaches and employs a single transformer backbone for joint feature extraction and matching. We show that SimVOS can greatly benefit from existing large-scale self-supervised pre-trained models (e.g., MAE) and can be served as a simple yet effective baseline for developing self-supervised pre-training tasks in VOS. Moreover, a new token refinement module is proposed to further reduce the computational cost and increase the inference speed of SimVOS. The proposed SimVOS achieves state-of-the-art performance on existing popular VOS benchmarks, \abc{and the simple design can inspire and serve as a baseline for future ViT-based VOS.}

\section{Acknowledgment}
This work was supported by a grant from the Research Grants Council of the Hong Kong Special Administrative Region, China (Project No. CityU 11215820), and a Strategic Research Grant from City University of Hong Kong (Project No. 7005840).

{\small
\bibliographystyle{ieee_fullname}
\bibliography{egbib}
}

\clearpage

\renewcommand\thesection{\Alph{section}}


In this supplementary material, we provide detailed implementation details, additional ablation study, completed comparison on YouTube-VOS 19, and more qualitative and quantitative results to demonstrate the effectiveness of the proposed \emph{Simplified VOS framework} (SimVOS). Specifically,  Sec.~\ref{arch} shows the detailed training details and architectures for our SimVOS. Sec.~\ref{speed} shows the speed comparison on V100 platform. More completed quantitative and qualitative results are respectively presented in Sec.~\ref{results} and Sec.~\ref{qualitative}.

\section{Implementation Details}
\label{arch}
\noindent\textbf{Training hyperparameters.} The training details of our SimVOS are shown in Tables \ref{davis_para} and \ref{davis_ytvos}. Following the previous VOS approaches \cite{STM,AOT,CFBI}, different training data sources are used to train our SimVOS model, which depends on the target evaluation benchmark. Specifically, for DAVIS 16/17 \cite{davis16,davis17} evalution,  the combination of the training splits in both DAVIS-17 \cite{davis17} and YouTube-VOS 19 \cite{youtubevos} is used for training. For YouTube-VOS 19 evaluation, only the training split in its own dataset is used. During the training stage, only a pair of frames is randomly sampled within the predefined maximum sampling frame gap. We use a larger maximum sampling frame gap (i.e., 15) for the YouTube VOS evaluation since the videos in this dataset are commonly longer than the videos in the DAVIS datasets. To alleviate overfitting and generalize well to unseen objects in YouTube-VOS, a larger droppath rate (i.e., 0.25) is employed for training.

\begin{table}[t]
  \newcommand{\tabincell}[2]
    \centering
    \begin{tabular}{c|c}
    \Xhline{\arrayrulewidth}
    \multicolumn{1}{c|}{Config}&  \multicolumn{1}{c}{Value}\cr
       \Xhline{\arrayrulewidth}  
    optimizer           &AdamW \cite{adam}   \cr  
  base learning rate           & 2e-5      \cr 
     weight decay             &1e-7    \cr  
     droppath rate           & 0.1 \cr
     batch size              &32       \cr 
            Iterations              &210,000       \cr 
       learning rate decay iteration              &125,000       \cr 
     learning rate schedule            & steplr     \cr  
     maximum sampling frame gap           & 10    \cr  
     training set & DAVIS \cite{davis17} + YT-VOS \cite{youtubevos}  \cr
   \Xhline{\arrayrulewidth}  
   \end{tabular}
   \vspace{-0.1cm}
   \centering
  \caption{The training parameters of SimVOS used for DAVIS \cite{davis16,davis17} evaluation.}  
  \label{davis_para}
\end{table}

   

\noindent\textbf{Architecture of the token refinement (TR) module.} The TR module consists of a convolutional layer and a fully-connected layer, which is denoted as $f(\cdot)$ in Eq. 3 of the main paper. The convolutional layer firstly reduces the input channel of $(C+1)$ to $C/4$ with a $3\times 3$ kernel, and the output is activated with a GELU \cite{gelu} function.  The fully-connected layer further maps the input channel of $c/4$ to $K$ for the following prototype generation, which is detail in Fig. 4 of the main paper.

\noindent\textbf{Training.} The training is conducted on 8 NVIDIA A100 GPUs, which takes about 15 hours to finish the whole main training on video datasets. 

\begin{table}[t]
  \newcommand{\tabincell}[2]
    \centering
    \begin{tabular}{c|c}
    \Xhline{\arrayrulewidth}
    \multicolumn{1}{c|}{Config}&  \multicolumn{1}{c}{Value}\cr
       \Xhline{\arrayrulewidth}  
    optimizer           &AdamW \cite{adam}   \cr  
  base learning rate           & 2e-5      \cr 
     weight decay             &1e-7    \cr  
     droppath rate           & 0.25 \cr
     batch size              &32       \cr 
            Iterations              &210,000       \cr 
       learning rate decay iteration              &125,000       \cr 
     learning rate schedule            & steplr     \cr  
     maximum sampling frame gap           &15    \cr  
     training set & YT-VOS \cite{youtubevos}  \cr
   \Xhline{\arrayrulewidth}  
   \end{tabular}
   \vspace{-0.1cm}
   \centering
  \caption{The training parameters of SimVOS  used for YouTube-VOS 19 \cite{youtubevos} evaluation.}  
  \label{davis_ytvos}
\end{table}

\begin{table*}[t]
  \newcommand{\tabincell}[2]
    \centering
    \begin{tabular}{ccccccc}
    \Xhline{\arrayrulewidth}
    \multirow{2}{*}{Method} & \multirow{2}{*}{S}&
  \multirow{2}{*}{$\mathcal{J}\&\mathcal{F}\uparrow$} &
  \multirow{2}{*}{$\mathcal{J}_{seen}\uparrow$} &
  \multirow{2}{*}{$\mathcal{F}_{seen}\uparrow$} &
   \multirow{2}{*}{$\mathcal{J}_{unseen}\uparrow$} &
    \multirow{2}{*}{$\mathcal{F}_{unseen}\uparrow$}\cr
  \cr
       \Xhline{\arrayrulewidth}
   MiVOS$^{*}$ \cite{bl30k}  & \cmark & 82.4 & 80.6 &84.7& 78.1 &86.4 \cr 
   HMMN \cite{HMMN}   & \cmark & 82.5 & 81.7 &86.1& 77.3 &85.0 \cr      
   STCN \cite{STCN}   & \cmark & 82.7 & 81.1 &85.4& 78.2&85.9\cr 
   STCN$^{*}$ \cite{STCN}   & \cmark & 84.2 & 82.6 &87.0& 79.4 & 87.7 \cr    
   SwinB-AOT \cite{AOT}   & \cmark & 84.5 & 84.0 &88.8& 78.4&86.7\cr  
   XMEM \cite{XMEM}   & \cmark & 85.5 & 84.3 &88.6& 80.3& 88.6 \cr	  
   XMEM$^{*}$ \cite{XMEM}   & \cmark & 85.8 & 84.8 &89.2& 80.3 &88.8 \cr	  
   \Xhline{\arrayrulewidth}
   CFBI \cite{CFBI}   &  & 81.0 & 80.6 &85.1& 75.2 & 83.0\cr          
   CFBI$+$ \cite{cfbi_plus}  & & 82.6 & 81.7 &86.2& 77.1 &85.2 \cr	
   JOINT \cite{joint} && \underline{82.8} & 80.8 &84.8& \underline{79.0} &86.6 \cr	
   SSTVOS \cite{SSTVOS}  & & 81.8 & 80.9 &-& 76.7&- \cr
   XMEM$^{-}$ \cite{XMEM}   &  & \textbf{84.2} & \textbf{83.8} &\textbf{88.3}& 78.1 &\underline{86.7} \cr	   
   \textbf{SimVOS-BS}   &  & {82.2} & {81.7} &{86.1}& {76.4} & {84.7}\cr	
   \textbf{SimVOS-B}   &  & \textbf{84.2} & \underline{83.1} &\underline{87.5}& \textbf{79.1} & \textbf{87.2}\cr		
   \Xhline{\arrayrulewidth}
   \end{tabular} 
   \vspace{-0.1cm}
   \centering
  \caption{ Comparisons with previous approaches on the YouTube-VOS 2019 validation set. S indicates the usage of synthetic data pre-training. $*$ denotes the BL30K \cite{bl30k} pretraining. $-$ means without applying synthetic data pre-training. We use the default 480P 6 FPS videos evaluation on YouTube-VOS 2019.} 
  \label{ytvos_compare}
\end{table*}

\begin{table*}[t]
  \newcommand{\tabincell}[2]
    \centering
     \resizebox{\linewidth}{!}{
  \footnotesize
     \begin{tabular}{@{}c@{\hspace{0.1cm}}@{\hspace{0.05cm}}|c@{\hspace{0.2cm}}cc|ccc|cccc@{}}
    \Xhline{\arrayrulewidth}
    \multirow{2}{*}{Method}& \multicolumn{3}{c|}{DAVIS-16}&\multicolumn{3}{c|}{DAVIS-17} & \multicolumn{3}{c}{YT-19} \cr
   & $\mathcal{J}\&\mathcal{F}$ & $\mathcal{J}$ & FPS & $\mathcal{J}\&\mathcal{F}$ & $\mathcal{J}$ & FPS & $\mathcal{J}\&\mathcal{F}$  & $\mathcal{J}_{unseen}$ & FPS\cr
       \Xhline{\arrayrulewidth} 
     
     STCN (NeurIPS`21) & - & - & 26.9& 82.5 & 79.3 & 20.2& - & - & 13.2 \cr
     SwinB-DeAOT-L (NeurIPS`22) & 89.8  & 88.7  & - & 83.8 & 81.0 & 15.4 & 82.0 & 76.1 & 11.9 \cr
     XMEM (ECCV`22) & 90.8 & 89.6 & 29.6 & 84.5 & 81.4 & 22.6 & \textbf{84.2} & 78.1 & 22.6 \cr
     SimVOS-BS     & 91.5 &  89.9 & 12.3  & 87.1 & 84.1 & 8.0 & 82.2 & 76.4 & 7.5 \cr
     SimVOS-B            &\textbf{92.9}& \textbf{91.3} & 7.2 & \textbf{88.0} & \textbf{85.0} & 3.5  & \textbf{84.2} & \textbf{79.1} & 3.3 \cr  
   \Xhline{\arrayrulewidth}  
   \end{tabular}}
   \centering
  \caption{Performance and FPS comparison between our SimVOS and SOTA approaches. All methods  use a \textbf{single  training stage} on DAVIS17+YT-19) for fair comparison. FPS is measured on one V100.
  }  
  \label{speed_accuracy_comp}
\end{table*}

\begin{figure*}
\begin{center}
   \includegraphics[width=1.0\linewidth]{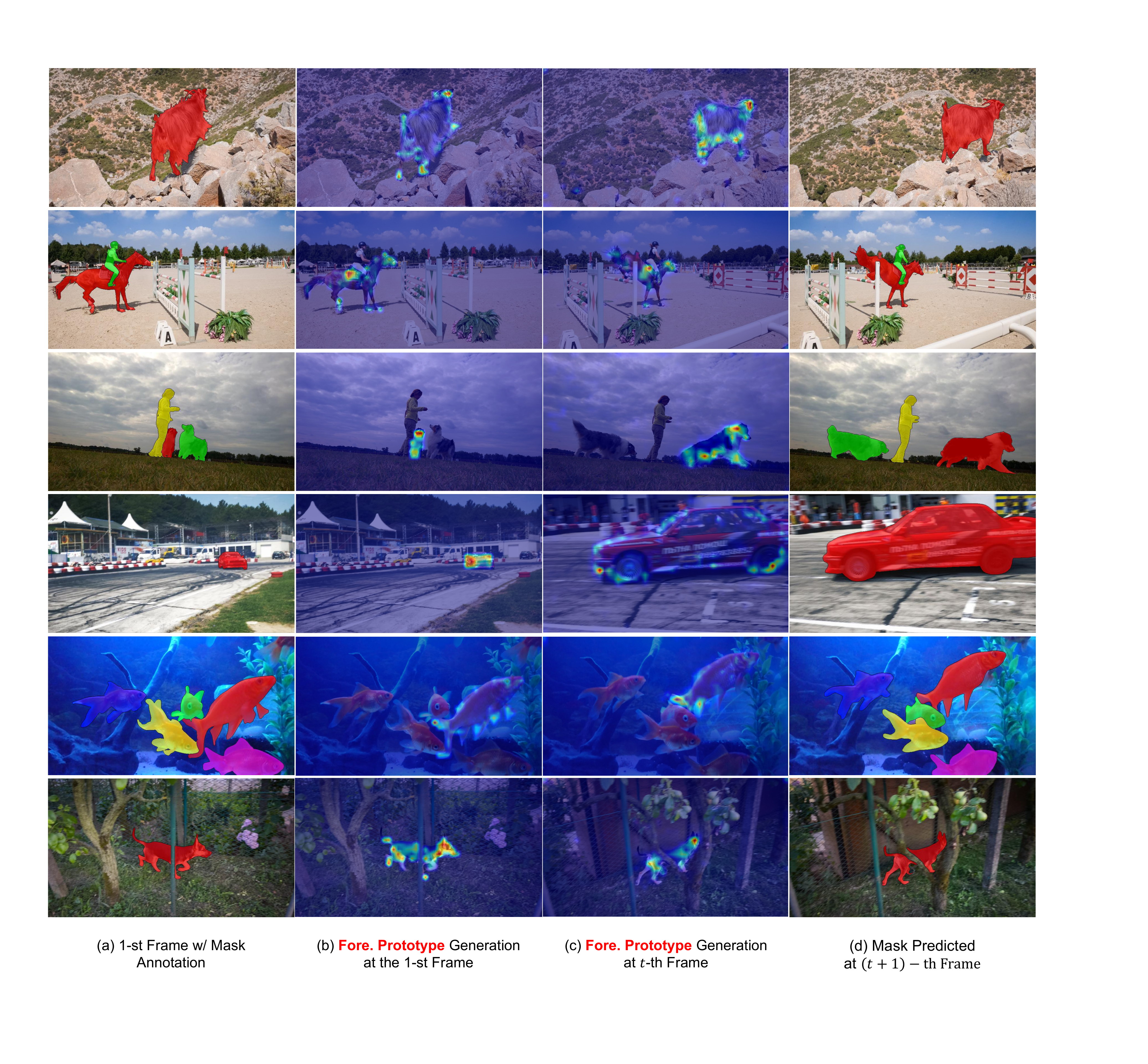} 
\end{center}
\vspace{-0.35cm}
 \caption{Visualization of (a) the mask annotation in the first frame, foreground prototype generation in both the (a) first frame and (b) previous frame (i.e., $t$-th frame), and the mask prediction in the following $(t+1)$-th frame. If multiple objects exist, the object annotated with \textcolor{red}{red mask} is chosen for visualization of the foreground prototype generation.}
 \label{supp_vis}
 \vspace{-0.35cm}
\end{figure*}


\section{Results on YouTube-VOS 19}
\label{results}
We show the complete results on YouTube-VOS 19 \cite{youtubevos} in Table \ref{ytvos_compare}. Our methods perform favorably against state-of-the-art VOS approaches under the same training setting. Specifically, SimVOS-B achieves better performance on unseen objects than the other approaches, which shows its generalization ability to new objects. Although our efficient variant (SimVOS-BS) obtains inferior results to SimVOS-B, it still outperforms the other transformer-based approach (SSTVOS) in terms of the $\mathcal{J}\&\mathcal{F}$ metric.

\section{Speed Comparison on the V100 platform}
\label{speed}
Tab.~\ref{speed_accuracy_comp} shows the speed using a V100 on 3 datasets.
Despite its lower speed, our SimVOS-B gets best performance on 3 VOS benchmarks w/ the naive memory mechanism, which demonstrates its strong matching ability. 
Our TR module reduces the generated tokens to speed-up VOS.
Other solutions are also possible, e.g., modifying ViT to be more efficient. We leave this as future work since our aim is to 
bridge the gap between VOS and self-supervised pre-training ViT communities, inspiring future works in VOS pre-training.

\section{Qualitative Visualization}
\label{qualitative}
We show more qualitative visualization in Fig. \ref{supp_vis}. The visualization of the attention in foreground prototype generation indicates that the TR module tends to aggregate discriminative boundary features. This can be explained that the local boundary cues play an essential role in accurate VOS. 

{\small
\bibliographystyle{ieee_fullname}
\bibliography{egbib}
}

\end{document}